\documentclass[10pt,twocolumn,letterpaper]{article}

\usepackage{cvpr}
\usepackage{times}
\usepackage{epsfig}
\usepackage{graphicx}
\usepackage{amsmath}
\usepackage{amssymb}

\usepackage[breaklinks=true,bookmarks=false]{hyperref}

\cvprfinalcopy 

\def\cvprPaperID{****} 

\begin{document}

\title{Multimodal Memory Modelling for Video Captioning}

\author{Junbo Wang$^1$  \hspace{7mm} Wei Wang$^1$  \hspace{7mm} Yan Huang$^1$ \hspace{7mm} Liang Wang$^{1,2}$ \hspace{7mm}  Tieniu Tan$^{1,2}$\\
$^1$Center for Research on Intelligent Perception and Computing\\
National Laboratory of Pattern Recognition\\
$^2$Center for Excellence in Brain Science and Intelligence Technology\\
Institute of Automation, Chinese Academy of Sciences\\
{\tt\small \{junbo.wang, wangwei, yhuang, wangliang, tnt\}@nlpr.ia.ac.cn}
}

\maketitle

\begin{abstract}
Video captioning which automatically translates video clips into natural language sentences is a very important task in computer vision. By virtue of recent deep learning technologies, e.g., convolutional neural networks (CNNs) and recurrent neural networks (RNNs), video captioning has made great progress. However, learning an effective mapping from visual sequence space to language space is still a challenging problem. In this paper, we propose a Multimodal Memory Model (M$^3$) to describe videos, which builds a visual and textual shared memory to model the long-term visual-textual dependency and further guide global visual attention on described targets. Specifically, the proposed M$^3$ attaches an external memory to store and retrieve both visual and textual contents by interacting with video and sentence with multiple read and write
operations. First, text representation in the Long Short-Term Memory (LSTM) based text decoder is written into the memory, and the memory contents will be read out to guide an attention to select related visual targets. Then, the selected visual information is written into the memory, which will be further read out to the text decoder. To evaluate the proposed model, we perform experiments on two publicly benchmark datasets: MSVD and MSR-VTT. The experimental results demonstrate that our method outperforms the state-of-the-art methods in terms of BLEU and METEOR.
\end{abstract}

\section{Introduction}

\begin{figure}[h]
\begin{center}
\includegraphics[scale=0.45]{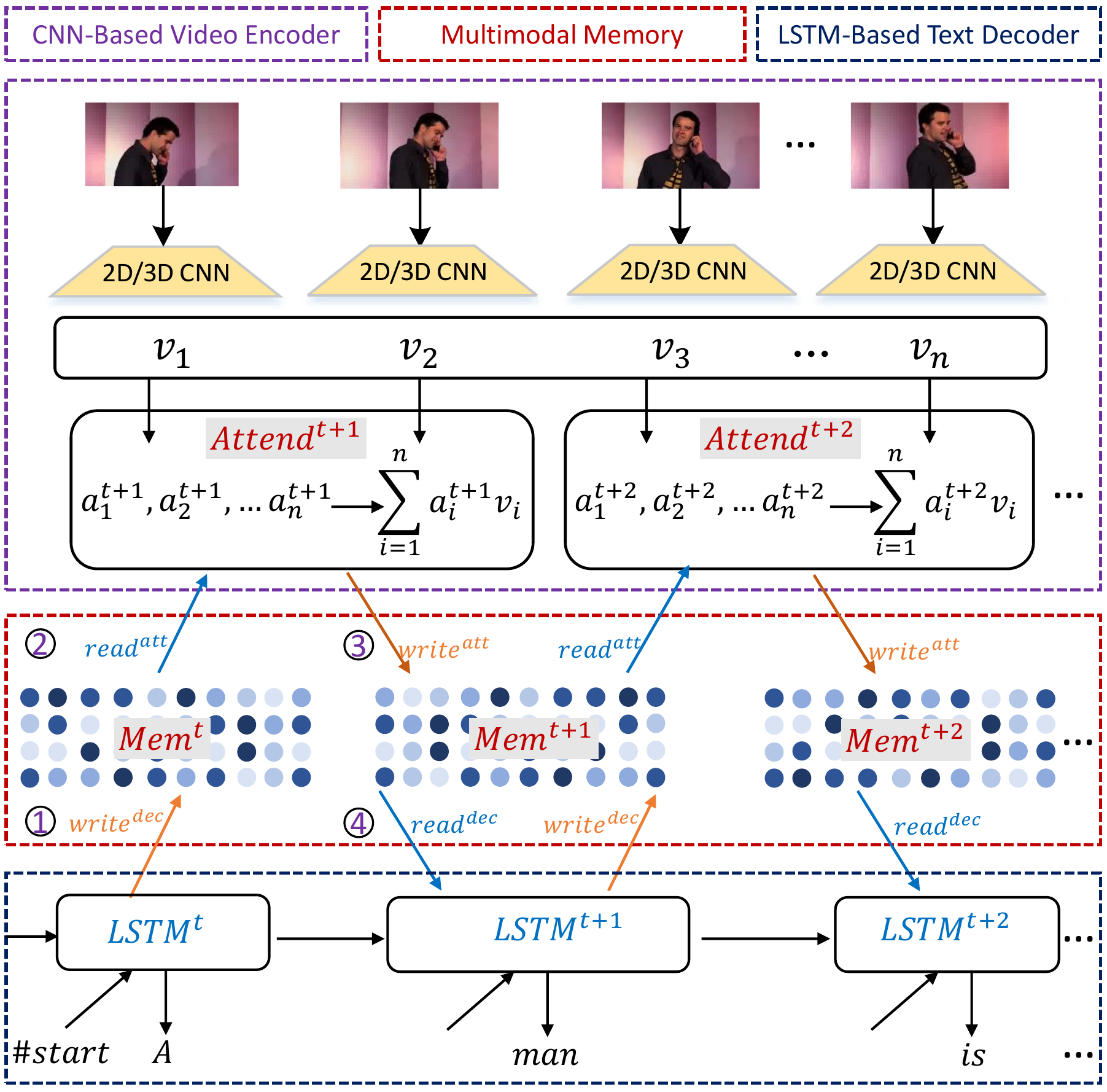}\\
\end{center}
\caption{The overall framework of multimodal memory modelling (M$^3$) for video captioning. It contains a CNN-based video encoder, a multimodal memory and a LSTM-based text decoder which are denoted by dashed box in different color. The multimodal memory $Mem$ stores and retrieves both visual and textual information by interacting with video and sentence with multiple read and write operations. The proposed M$^3$ with explicit memory modelling can not only model the long-term visual-textual dependency, but also guide global visual attention for effective video representation. (Best viewed in color)}
\label{figure:architect}
\end{figure}


Describing videos with natural sentences automatically also called video captioning is very important for bridging vision and language, which is also a very challenging problem in computer vision. It has plenty of practical applications, e.g., human-robot interaction, video indexing and describing videos for the visually impaired.

Video captioning involves in understanding both vision and language, and then builds the mapping from visual elements to words. As we know, video as image sequence contains rich information about actor, object, action, scene and their interactions. It is very difficult for existing methods using a single visual representation \cite{venugopalan2014translating} to capture all these information over a long period. Yao et al. \cite{yao2015describing} attempt to dynamically select multiple visual representations based on temporal attention mechanism which is driven by the hidden representations from a Long Short-Term Memory (LSTM) text decoder. The LSTM text decoder, which integrates the information from both words and selected visual contents, models the sentence generation and guides visual selection. However, recent work \cite{graves2014neural} has pointed out that LSTM doesn't work well when the sequence is long enough. Current video captioning benchmark datasets generally have long sentences to describe videos. The consequence is that the LSTM-based decoder can not model the long-term visual-textual dependency well and can not guide visual attention to select the described targets.

Recently, neural memory models have been proposed and successfully applied to question answering \cite{xiong2016dynamic} and dialog systems \cite{dodge2015evaluating}, which pose potential advantages to long-term dependency modelling in sequential problems. In addition, as Wang et al. \cite{wang2011simulating} said, visual working memory is one of the key factors to guide eye movements. That is to say, explicitly introducing memory into video captioning can not only model the long-term visual-textual dependency, but also guide visual attention for better video representation.

%

In this paper, we propose a Multimodal Memory Model (M$^3$) to describe videos, which builds a visual and textual shared memory to model the long-term visual-textual dependency and further guide global visual attention on described targets. Similar to Neural Turing Machines \cite{graves2014neural}, the proposed M$^3$ attaches an external memory to store and retrieve both visual and textual information by interacting with video and sentence with multiple read and write operations. Fig. 1 shows the overall framework of multimodal memory modelling for video captioning, which consists of three key components: convolutional neural networks (CNN) based video encoder, multimodal memory and Long Short-Term Memory (LSTM) based text decoder. (1) CNN-based video encoder first extracts video frame/clip features using pretrained 2D/3D CNNs which are often used for image/video classification. The extracted features $\{v_i\}_{i=1}^n$ form the original video representation. Similar to \cite{yao2015describing}, temporal soft-attention $Attend$ is used to select most related visual information to each word. But very different from \cite{yao2015describing} using the hidden states from a LSTM decoder, we guide the soft-attention based on the content from a multimodal memory ($read^{att}$ in Fig.1 denotes the content read from memory for attention). Then the selected visual information will be written into the memory ($write^{att}$ denotes the content written to memory from selective attention). (2) LSTM-based text decoder models the sentence generation with a LSTM-RNN architecture, which predicts the $\{t+1\}^{th}$ word conditioned on not only previous hidden representation $LSTM^t$ but also the content read from the multimodal memory ($read^{dec}$ denotes the content read from memory for decoder). Besides word prediction, the text decoder also writes the updated representation to the memory ($write^{dec}$ denotes the content written to memory from the decoder). (3) Multimodal memory contains a memory matrix $Mem$ to interact with video and sentence, e.g., write hidden representation from the LSTM decoder to memory $write^{dec}$, read memory contents for the decoder $read^{dec}$. Each write operation will update the multimodal memory, e.g., from $Mem^t$ to $Mem^{t+1}$. In Fig.1, we illustrate the procedure of memory-video/sentence interactions: \textcircled{1} write hidden states to update memory, \textcircled{2} read the updated memory content to perform soft-attention, \textcircled{3} write selected visual information to update memory again, \textcircled{4} read the updated memory content for next word prediction.

We evaluate our model on two publicly benchmark datasets, e.g., Microsoft Research
Video Description Corpus (MSVD) and Microsoft Research-Video to Text (MSR-VTT). The proposed M$^3$ achieves the state-of-the-art performance, which demonstrates the effectiveness of our model.
\section{Related Work}
In this section, we briefly introduce some existing work that closely related to our proposed model.

{\bf Video Captioning} \hspace{3mm} Video captioning has been researched for a long period due to its importance in bridging vision and language. Various methods have been proposed to solve this problem, which can be categorized into three classes. The first class \cite{guadarrama2013youtube2text,krishnamoorthy2013generating,thomason2014integrating} detect the attributes of given videos and derive the sentence structure with predefined sentence templates. Then probabilistic graphical models are used to align the phases to the attributes. Similar to image captioning, this kind of methods always generate grammatically correct sentences, but lose the novelty and flexibility of the sentence. The second class of methods \cite{huang2012multi,wei2010multimodal} treat video captioning as a retrieval task, which retrieve similar videos from external databases and recompose the descriptions of the retrieved videos to gain target sentence. This kind of methods generally generate more natural sentences than the first class, but have a strong dependency upon external databases.
The third class of methods inspired by Neural Machine Translation (NMT) \cite{kalchbrenner2013recurrent,cho2014learning} map video sequence to sentence by virtue of recent deep neural networks, e.g., CNNs and RNNs. Venugopalan et al. \cite{venugopalan2014translating} apply average pooling to extract the features of multiple video frames and use a two-layer LSTM network on these features to generate descriptions. In order to enhance video representation, Ballas et al. \cite{ballas2015delving} exploit the intermediate visual representation extracted from pre-trained image classification models, and Pan et al. \cite{pan2015hierarchical} propose a hierarchical recurrent neural encoder to explore the temporal transitions with different granularities. In order to generate more sentences for each video, Yu et al. \cite{yu2015video} exploit a hierarchical recurrent neural network decoder which contains a sentence generator and a paragraph generator. To emphasize the mapping from video to sentence, Yao et al. \cite{yao2015describing} propose a temporal attention model to align the most relevant visual segments to the generated captions, and Pan et al. \cite{pan2015jointly} propose a long short-term memory with a visual-semantic embedding model. Recently, the third class of deep learning based methods have made great progress in video captioning. We augment existing deep learning based models with an external memory to model the long-term visual-textual dependency and guide global visual attention in this paper.

{\bf Memory Modelling} \hspace{3mm} To extend the memory ability of traditional neural networks, Graves et al. \cite{graves2014neural} propose a Neural Turing Machine (NTM) which holds an external memory to interact with the internal state of neural networks by attention mechanism. NTM has shown the potential of storage and access of information over long time periods which has always been problematic for RNNs, e.g., copying, sorting and associative recall. Besides memory matrix in NTM, memory is also modelled as continuous and differentiable doubly-linked lists and stacks \cite{joulin2015inferring}, queues and deques \cite{grefenstette2015learning}. Different from exploring various forms of dynamic storages,  Weston et al. \cite{weston2014memory} model large long-term static memory. The internal information stored in the static memory is not modified by external controllers, which is specially used for reading comprehension. These memory networks have been successfully applied to the tasks which need long-term dependency modelling, e.g., textual question answering \cite{bordes2015large, hermann2015teaching}, visual question answering \cite{xiong2016dynamic} and dialog systems \cite{dodge2015evaluating}. As we know, few memory models have been proposed for video captioning. In this paper, we will propose an external multimodal memory to interact with video and sentence simultaneously.


\section{Multimodal Memory Modelling for Video Captioning}\label{sect:model}
In this section, we will first introduce three key components of our model including: 1) convolutional neural networks (CNN) based video encoder, 2) Long Short-Term Memory (LSTM) based text decoder, and 3) multimodal memory. Then we will explain the procedure of model training and inference in details.

\subsection{CNN-Based Video Encoder}

Convolutional neural networks (CNNs) have achieved great success in many computer vision tasks recently, e.g., image classification \cite{krizhevsky2012imagenet} and object detection \cite{girshick2014rich}. Due to the power of representation learning, CNNs pre-trained by these tasks can be directly transferred to other computer vision tasks as generic feature extractors.

In our model, we consider using pre-trained 2D CNNs to extract appearance features of videos,
and pre-trained 3D CNNs to obtain motion features of videos since the temporal dynamics is very important for video understanding.
In particular for an input video, we first sample it with fixed number of frames/clips,
and then exploit the pre-trained 2D CNNs/3D CNNs to extract features of each frame/clip.
We denote the obtained video representation as $V = \{ {v_1},{v_2},{v_3}, \ldots ,{v_n}\}$, where $n$ is the number of sampled frames/clips.


\subsection{LSTM-Based Text Decoder}

Similar to previous work \cite{bahdanau2014neural,vinyals2015show,venugopalan2014translating},
we use a LSTM network \cite{hochreiter1997long} as our language model to model the syntactic structure of sentence,
which can deal with input word sequences of arbitrary length by simply setting the start and end tags.
The LSTM network has the similar architecture as a standard RNN,
except for the hidden unit is replaced by a LSTM memory cell unit.
But better than standard RNNs, the LSTM network can considerably release the gradient vanishing problem,
which has thus accomplished better performance in natural language processing applications \cite{bahdanau2014neural}.

Although LSTM has many variants, here we use a widely used one described in \cite{zaremba2014recurrent}.
It includes a single memory cell, an input activation function, an output activation function
and three gates (i.e., input gate, forget gate and output gate).
The memory cell ${c_t}$ records the history of all observed inputs up to the current time,
by recurrently summarizing the previous memory cell ${c_{t-1}}$
and the candidate cell state $\tilde{\text{c}}_t$,
modulated by a forget gate ${f_t}$ and an input gate ${i_t}$, respectively.
The input gate ${i_t}$ utilizes the input to change the state of the memory cell ${c_t}$, the forget gate ${f_t}$ allows the memory cell to adaptively
remember or forget its previous state, and the output gate ${o_t}$ modulates the state of memory cell ${c_t}$ to output the hidden state.

Different from the commonly used unimodal LSTM,
we incorporate the fused multimodal information ${r_t}$ as another input,
which is read from our multimodal memory during caption generation as demonstrated in next section.
For given sentences, we use one-hot vector encoding to represent each word.
By denoting the input word sequence as $\left\{ {{y_t}|t = 0,1,2,\cdots,T} \right\}$,
and the corresponding embedding vector of word ${y_t}$ as ${E_t}$,
the hidden activations $h_t$ at time $t\left( {t = 1, \cdots ,T } \right)$
can be computed as follows.
\begin{equation}
{i_t} = \sigma \left( {{W_i}{E_{t - 1}} + {U_i}{h_{t - 1}} + {M_i}{r_t} + {b_i}} \right)
\end{equation}
\begin{equation}
{f_t} = \sigma \left( {{W_f}{E_{t - 1}} + {U_f}{h_{t - 1}} + {M_f}{r_t} + {b_f}} \right)
\end{equation}
\begin{equation} {o_t} = \sigma \left( {{W_o}{E_{t - 1}} + {U_o}{h_{t - 1}} + {M_o}{r_t} + {b_o}} \right)
\end{equation}
\begin{equation}
{\tilde{\text{c}}_t} = \phi \left( {{W_c}{E_{t - 1}} + {U_c}{h_{t - 1}} + {M_c}{r_t} + {b_c}} \right)
\end{equation}
\begin{equation}
{c_t} = {i_t} \odot \tilde{\text{c}}_t  + {f_t} \odot {c_{t - 1}}
\end{equation}
\begin{equation}
{h_t} = {o_t} \odot \phi \left( {{c_t}} \right)
\end{equation} where the default operation between matrices is matrix multiplication, $\odot$ denotes an element-wise multiplication, $W$, $U$, and $M$ denote the shared weight matrices to be learned, and $b$ denotes the bias term. $\tilde{\text{c}}_t$ is the input to the memory cell ${c_t}$, which is gated by the input gate ${i_t}$. $\sigma$ denotes the element-wise logistic sigmoid function, and $\phi$ denotes hyperbolic tangent function tanh.

For clear illustration, the process of language modelling mentioned above can be abbreviated as follows.
\begin{equation}
{h_t} = \psi \left( {{h_{t - 1}},{c_{t - 1}},{y_{t - 1}},{r_t}} \right)
\end{equation}

\subsection{Multimodal Memory}
Although LSTM network can well address the ``the vanishing and exploding gradient'' problem \cite{hochreiter2001gradient},
it cannot deal with very long sequences due to the limited capacity of memory cells.
Considering that Neural Turing Machine (NTM) \cite{graves2014neural} can capture very long
rang temporal dependency with external memory,
we attach a shared multimodal memory between the LSTM-based language model and CNN-based visual model
for long range visual-textual information interaction.

Our multimodal memory at time t is a $N \times M$ matrix ${M_t}$,
where N denotes the number of memory locations and M denotes the vector length of each location.
The memory interacts with the LSTM-based language model and CNN-based visual model via selective read and write operations.
Since there exists bimodal information, i.e., video and language, we employ two independent read/write operations to guide the information interaction.
As shown in Fig. \ref{figure:architect}, during each step of sentence prediction, the hidden representation in LSTM-based language model is first written into the multimodal memory, and the memory contents will be read out to guide a visual attention scheme to select relevant visual information.
Then, the selected visual information is written into the memory, which will be further read out for language modelling. In the following, we will introduce the details of this procedure.

\subsubsection{Memory Interaction}
The interaction of visual information and textual elements is performed in the following order.

{\bf Writing hidden representations to update memory} \hspace{3mm} Before predicting the next word during the process of caption generation, our LSTM-based language model will write previous hidden representations into the multimodal memory, to summarize the previous textual information. We denote the current textual weighting vector, textual erase vector and textual add vector as ${w_t^{sw}}$, ${e_t^{sw}}$ and ${a_t^{sw}}$, respectively, all of which are emitted by the LSTM-based language model. The elements of textual erase vector ${e_t^{sw}}$ lie in the range (0,1). The lengths of textual erase vector ${e_t^{sw}}$ and textual add vector ${a_t^{sw}}$ are both M. Since both the textual erase vector and textual add vector have M independent elements, the elements in every memory location can be erased or added in a fine-grained way. Then the textual information can be written into the memory as follows.
\begin{equation}
{M_t}\left( i \right) = {M_{t - 1}}\left( i \right)\left[ {1 - w_t^{sw}\left( i \right)e_t^{sw}} \right] + w_t^{sw}\left( i \right)a_t^{sw}
\end{equation}

{\bf Reading the updated memory for temporal attention} \hspace{3mm} After writing textual information into the memory, the updated memory content is read out to guide a visual attention model to select prediction-related visual information. Assuming that the current visual weighting vector over the N locations at time t is ${w_t^{vr}}$, which needs to be normalized as follows.
\begin{equation}
\sum\limits_{i = 1}^N {w_t^{vr}\left( i \right)}  = 1,{\rm{      0}} \le w_t^{vr}\left( i \right) \le 1,{\rm{ }}\forall {\rm{i}} \in \left[ {1,N} \right]
\end{equation}
Then the visual read vector ${r_t^{vr}}$ returned by the visual attention model is computed as a linear weighting of the row-vectors ${M_t}\left( i \right)$:
\begin{equation}
r_t^{vr} = \sum\limits_{i = 1}^N {w_t^{vr}\left( i \right)} {M_t}\left( i \right)
\end{equation}

{\bf Temporal attention selection for video representation} \hspace{3mm}
After reading the updated memory content, we use it for our visual attention selection. In particular, for the features $V = \{ {v_1},{v_2},{v_3}, \ldots ,{v_n}\}$ of video frames/clips, instead of directly feeding them to the latter LSTM decoder by simple average pooling \cite{venugopalan2014translating}, we apply a soft attention mechanism \cite{yao2015describing} to select most relevant appearance and motion features while
largely preserving the temporal dynamics of video representation.
Taking a video sequence for illustration, the attention model aims to focus on specific object and action at one time. Specially, the attention model first computes the unnormalized relevance scores between the i-th temporal feature and current content read from the multi-modal memory, which summarizes embedding information of all the previously generated words, i.e., ${y_1},{y_2}, \ldots ,{y_{t - 1}}$. The unnormalized relevance score can be represented as follows.
\begin{equation}
e_i^t = {w^{\rm{T}}}\tanh \left( {{W_r}{r_t^{vr}} + {U_\alpha }{v_i} + {b_\alpha }} \right)
\end{equation}
where ${{W_r}}$, ${{U_\alpha }}$, ${{b_\alpha }}$, and $w$ are the parameters that are learned together with all other modules in the training network. Different from \cite{yao2015describing}, here we incorporate the content read from multimodal memory instead of the previous hidden state from LSTM network. We argue that the hidden state from LSTM network can not fully represent all the information of previous words, while our multimodal memory can well keep them.
After computing unnormalized relevance scores $e_i^t\left( {i = 1,2, \cdots ,n} \right)$ for all the frames in the input video, we use a softmax layer to obtain the normalized attention weights $\alpha _i^t$:
\begin{equation}
\alpha _i^t = \exp \left\{ {e_i^t} \right\}/\sum\limits_{j = 1}^n {\exp \left\{ {e_j^t} \right\}}
\end{equation}
Finally, we sum all the products of element-wise multiplication between the attention weights and appearance and motion features
to get the final representation of input video:
\begin{equation}
{V_t} = \sum\limits_{i = 1}^n {\alpha _i^t{v_i}}
\end{equation}

The above attention model allows the LSTM-based language model to selectively focus on the specific frames by increasing corresponding weights, which is very effective when there exist explict visual-semantic mappings. However, when the high-level visual representation is not intrinsically relevant to the word in the generated sentence, e.g. ``the'', ``number'' and ``from'', the predicting word is not relevant to any frame feature in the input video. In this case, the attentive video representation could act as the noise for the LSTM-based language model. To avoid this issue, we append a blank feature whose values are all zeros to the video features along the temporal dimension. Therefore, we can keep the sum of attention weights less equal to one:
\begin{equation}
\sum\limits_{i = 1}^n {\alpha _i^t}  \le 1
\end{equation}

{\bf Writing selected visual information to update memory} \hspace{3mm}
After selecting visual information via the attention model above, the information will be written into the memory for updating. Similar to the operation of writing hidden representations into the memory, the current visual weighting vector ${w_t^{vw}}$, visual erase vector ${e_t^{vw}}$ and visual add vector ${a_t^{vw}}$ are all emitted by the visual attention model. The elements of visual erase vector ${e_t^{vw}}$ lie in the range (0,1). The lengths of visual erase vector ${e_t^{vw}}$ and visual add vector ${a_t^{vw}}$ are both M. Then the visual information can be written into the memory as follows.
\begin{equation}
{M_t}\left( i \right) = {M_t}\left( i \right)\left[ {1 - w_t^{vw}\left( i \right)e_t^{vw}} \right] + w_t^{vw}\left( i \right)a_t^{vw}
\end{equation}

{\bf Reading the updated memory for LSTM-based language model} \hspace{3mm}
When finishing the above writing operation, the updated memory is read out for language
modelling. Similarly, assuming that the textual weighting vector over the N locations at the current time is ${w_t^{sr}}$, which also has to be normalized as follows.
\begin{equation}
\sum\limits_{i = 1}^N {w_t^{sr}\left( i \right)}  = 1,{\rm{      0}} \le w_t^{sr}\left( i \right) \le 1,{\rm{ }}\forall {\rm{i}} \in \left[ {1,N} \right]
\end{equation}
Then the textual read vector ${r_t^{sr}}$ returned by the LSTM-based language model is computed as a linear weighting of the row-vectors ${M_t}\left( i \right)$:
\begin{equation}
r_t^{sr} = \sum\limits_{i = 1}^N {w_t^{sr}\left( i \right)} {M_t}\left( i \right)
\end{equation}

{\bf Computing of RNN-based language model} \hspace{3mm} After getting the reading information from the updated memory, we can compute the current hidden state of LSTM-based language model by calling the following function.
\begin{equation}
{h_t} = \psi \left( {{h_{t - 1}},{c_{t - 1}},{y_{t - 1}},{r_t^{sr}}} \right)
\end{equation}

\subsubsection{Memory Addressing Mechanisms}
As stated in \cite{sukhbaatar2015end,graves2014neural}, the objective function is hard to converge when using a location-based addressing strategy.
Therefore, we use a content-based addressing strategy to update the above read/write weighting vector. During the process of content-based addressing, each read/write head first produces a key vector ${k_t}$ and a sharpening factor ${\beta _t}$. The key vector ${k_t}$ is mainly used for comparing with each memory vector ${M_t}\left( i \right)$ by a similarity measure function $K$, and the sharpening factor ${\beta _t}$ is employed for regulating the precision of the focus. Then all of them can be computed as follows.
\begin{equation}
{k_t} = \partial \left( {{w_k}{h_t} + {b_k}} \right)
\end{equation}
\begin{equation}
{\beta _t} = \delta \left( {{w_\beta }{h_t} + {b_\beta }} \right)
\end{equation}
\begin{equation}
K\left( {x,y} \right) = \frac{{x \cdot y}}{{\left\| x \right\| \cdot \left\| y \right\| + \varepsilon }}
\end{equation}
\begin{equation}
{z_t}\left( i \right) = {\beta _t}K\left( {{k_t},{M_t}\left( i \right)} \right)
\end{equation}
\begin{equation}
{w_t}\left( i \right) = \frac{{\exp \left( {{z_t}\left( i \right)} \right)}}{{\sum\limits_{j = 1}^N {\exp \left( {{z_t}\left( j \right)} \right)} }}
\end{equation}


\subsection{Training and Inference}
Assuming that there are totally N training video-description pairs $\left( {{x^i},{y^i}} \right)$ in the entire training dataset,
 where the description ${{y^i}}$ has a length of ${t_i}$. The overall objective function used in our model is the averaged log-likelihood over the whole training dataset plus a regularization term.
\begin{equation}
L\left( \theta  \right) = \frac{1}{N}\sum\limits_{i = 1}^N {\sum\limits_{j = 1}^{{t_i}} {\log \rho \left( {y_j^i|y_{1:j - 1}^i,{x^i},\theta } \right)} }  + \lambda \left\| \theta  \right\|_2^2
\end{equation}
where ${y_j^i}$ is a one-hot vector used to denote the input word, $\theta$ is all parameters to be optimized in the model, and $\lambda$ denotes the regularization coefficient. As all components in our model including multimodal memory components are differential, we can use Stochastic Gradient Descent (SGD) to learn the parameters.

Similar to most LSTM language models, we use a softmax layer to model the next word's probability distribution over the whole vocabulary.
\begin{equation}
{z_t} = tanh\left( {{W_v}{V_t} + {W_h}{h_t} + {W_e}{y_{t - 1}} + {b_h}} \right)
\end{equation}
\begin{equation}
{\rho _t} = softmax\left( {{U_\rho }{z_t} + {b_\rho }} \right)
\end{equation}
where ${{W_v}}$,${{W_h}}$,${{W_e}}$,${{b_h}}$,${{U_\rho }}$, and ${{b_\rho }}$ are the parameters to be estimated. Based on the probability distribution ${\rho _t}$, we can recursively sample ${y_t}$ until obtaining the end of symbol in the vocabulary.


During caption generation, we could directly choose the word with maximum probability at each timestep. However, the resulting generated sentences usually have low quality due to the local optimum strategy. Ideally, we should traverse all possible word at each timestep during the caption generation. But the exhausted search has very high computational cost, so we choose a beam search strategy to generate the caption, which is a fast and effective method \cite{yu2015video}.



\section{Experiments}

To validate the effectiveness of the proposed model, we do extensive experiments on two public video captioning datasets. The one is Microsoft Video Description Dataset (MSVD) \cite{chen2011collecting} which is tested by most of the state-of-the-art methods. The other is recently released Microsoft Research-Video to Text (MSR-VTT) \cite{Xu2016large} which is the largest dataset in terms of sentence and vocabulary.

\subsection{Datasets}
{\bf Microsoft Video Description Dataset} \hspace{3mm} Microsoft Video Description Dataset (MSVD) \cite{chen2011collecting} consists of 1970 videos which range from 10 seconds to 25 seconds. Each video has multi-lingual descriptions which are labelled by the Amazon's Mechanical Turk workers. For each video, the descriptions depict a single activity scene with about 40 sentences. So there are about 80,000 video-description pairs. Following the standard split \cite{yao2015describing,pan2015jointly}, we divide the original dataset into a training set of 1200 videos, a validation set of 100 videos, and a test set of 670 videos, respectively.

{\bf Microsoft Research-Video to Text Dataset} \hspace{3mm} Microsoft Research-Video to Text Dataset (MSR-VTT) is the recently released largest dataset in terms of sentence and vocabulary, which consists of 10,000 video clips and 200,000 sentences. Each video clip is labelled with about 20 sentences. Similar to MSVD, the sentences are annotated by Amazon's Mechanical Turk workers. With the split in \cite{Xu2016large}, we divide the original dataset into a training set of 6513 videos, a validation set of 497 videos and a testing set of 2990 videos, respectively.
\subsection{Data Preprocessing}

{\bf Video Preprocessing} \hspace{3mm} Instead of extracting features for each video frame, we uniformly sample $K$ frames from original video for feature extraction. When the video length is less than $K$, we pad zero frames at the end of original frames. Empirically, we set $K$ to 28 for 98 frames per video in MSVD, and set $K$ to 40 for 149 frames per video in MSR-VTT. For the extensive comparisons, we extract features from both pretrained 2D CNN networks, e.g., GoogleNet \cite{szegedy2015going}, VGG-19 \cite{simonyan2014very}, Inception-V3 \cite{szegedy2015rethinking}, ResNet-50 \cite{he2015deep}, and 3D CNN networks, e.g., C3D \cite{tran2015learning}. Specifically, we extract the features of the pool5/7x7$\_$s1 layer in GoogleNet, the fc7 layer in VGG-19, the pool3 layer in Inception-V3, the pool5 layer in ResNet-50 and the fc6 layer in C3D. 

{\bf Description Preprocessing} \hspace{3mm} The descriptions in MSVD and MSR-VTT are all converted into lower case. To reduce unrelated symbols, we tokenize all sentences by NLTK toolbox \footnote{http://www.nltk.org/index.html} and remove punctuations. The vocabulary in MSVD is about 13,000 while the vocabulary in MSR-VTT is about 29,000. For convenience, we set the vocabulary size to 20,000 for both datasets. So the rare words in MSR-VTT are eliminated to further reduce the vocabulary.

\subsection{Evaluation Metrics}
In this paper, we adopt two standard evaluation metrics: BLEU \cite{papineni2002bleu} and METEOR \cite{denkowski2014meteor}, which are widely used in machine translation and image/video captioning. The BLEU metric measures the n-grams precision between generated sentence and original description, which correlates highly with human evaluation results. The METEOR metric measures the word correspondences between generated sentences and reference sentences by producing an alignment \cite{chen2015microsoft}. METEOR is often used as a supplement to BLEU. To guarantee a fair comparison with previous methods, we utilize the Microsoft COCO Caption Evaluation tool \cite{chen2015microsoft} to gain all experimental results.

\subsection{Experimental Settings}

During model training, we add a start tag and an end tag to the sentence in order to deal with variable-length sentences. We also add masks to both sentences and visual features for the convenience of batch training. Similar to \cite{yao2015describing}, the sentences with length larger than 30 in MSVD and the sentences with length larger than 50 in MSR-VTT are removed. For the unseen words in the vocabulary, we set them to unknown flags. Several other parameters, e.g., word embedding dimension (468), beam size (5) and the size of multimodal memory matrix (128,512), are set using the validation set. To reduce the overfitting during training, we apply dropout \cite{srivastava2014dropout} with rate of 0.5 on the output of fully connected layers and the output of LSTMs but not on the recurrent transitions. To further prevent gradient explosion, we clip the gradients to [-10,10]. The optimization algorithm is ADADELTA \cite{zeiler2012adadelta} which we find fast in convergence.

\subsection{Experimental Results}

\begin{table}[h]\footnotesize
\addtolength{\tabcolsep}{-2pt}
\centering
\begin{tabular}{lccccc}

\hline
\hline
Method     & B@1    & B@2    & B@3    & B@4    & METEOR     \\
\hline
\hline
FGM \cite{thomason2014integrating}      &  -          &  -         &  -         &  13.68\%   &  23.90\% \\
LSTM-YT \cite{venugopalan2014translating}  &  -          &  -         &  -         &  33.29\%    &  29.07\% \\
SA \cite{yao2015describing}       &  -          &  -         &  -         &  40.28\%    &  29.00\% \\
S2VT \cite{venugopalan2015sequence}     &  -          &  -         &  -         &  -         &  29.2\% \\
LSTM-E \cite{pan2015jointly}  &  74.9\%     &  60.9\%    &  50.6\%    &  40.2\%    &  29.5\% \\
p-RNN \cite{yu2015video}    &  77.3\%     &  64.5\%    &  54.6\%    &  44.3\%    &  31.1\% \\
HRNE \cite{pan2015hierarchical}     &  79.2\%     &  66.3\%    &  55.1\%    &  43.8\%    &  \bf{33.1\%} \\
BGRCN \cite{ballas2015delving}   &  -          &  -         &  -         &  49.63\%   &  31.7\% \\
\hline
\hline
M$^3$-c3d        &  77.30\%     &  68.20\%    &  56.30\%    &  45.50\%   &  29.91\% \\
M$^3$-vgg19      &  77.70\%     &  67.50\%    &  58.90\%    &  49.60\%   &  30.09\% \\
M$^3$-google     &  79.05\%     &  68.74\%    &  60.00\%    &  51.17\%   &  31.47\% \\
M$^3$-res        &  80.80\%     &  69.90\%    &  60.40\%    &  49.32\%   &  31.10\% \\
M$^3$-inv3       &  \bf{81.56\%}     &  \bf{71.39\%}    &  \bf{62.34\%}    &  \bf{52.02\%}   &  32.18\% \\
\hline
\hline
\end{tabular}
\vspace{3mm}
\caption{The performance comparison with the other eight state-of-the-art methods using single visual feature on MSVD. The results of the proposed M$^3$ with five single features are shown at the bottom of the table. We compare the best single feature results of the other eight methods at the top of the table.}
\label{msvd-rgb}
\end{table}

\begin{table}[h]\footnotesize
\addtolength{\tabcolsep}{-3pt}
\centering
\begin{tabular}{lccccc}

\hline
\hline
Method     & B@1    & B@2    & B@3    & B@4    & METEOR     \\
\hline
\hline
SA-G-3C \cite{yao2015describing}  &  -          &  -         &  -         &  41.92\%    &  29.60\% \\
S2VT-rgb-flow \cite{venugopalan2015sequence}  &  -      &  -         &  -         &  -         &  29.8\% \\
LSTM-E-VC \cite{pan2015jointly}   &  78.8\%     &  66.0\%    &  55.4\%    &  45.3\%    &  31.0\% \\
p-RNN-VC \cite{yu2015video}   &  81.5\%     &  70.4\%    &  60.4\%    &  49.9\%    &  32.6\% \\

\hline
\hline
M$^3$-VC       &  81.90\%    &  71.26\%     &  62.08\%    &  51.78\%    &  32.49\% \\
M$^3$-IC       &  \bf{82.45\%}    &  \bf{72.43\%}     &  \bf{62.78\%}    &  \bf{52.82\%}    &  \bf{33.31\%} \\ \hline
\hline
\end{tabular}
\vspace{3mm}
\caption{The performance comparison with the other four state-of-the-art methods using multiple visual feature fusion on MSVD. Here V, C, I and G denote VGG-19 \cite{simonyan2014very}, C3D \cite{tran2015learning}, Inception-V3 \cite{szegedy2015rethinking} and GoogleNet \cite{szegedy2015going}, respectively.}
\label{msvd-fuse}
\end{table}

\begin{table}[h]\footnotesize
\addtolength{\tabcolsep}{-2pt}
\centering
\begin{tabular}{lccccc}
\hline
\hline
Method     & B@1    & B@2    & B@3    & B@4    & METEOR     \\
\hline
\hline
SA-V \cite{yao2015describing}      &  67.82\%     &  55.41\%    &  42.90\%    &  34.73\%    &  23.11\% \\
SA-C \cite{yao2015describing}      &  68.90\%     &  57.50\%    &  47.00\%    &  37.40\%    &  24.80\% \\
SA-VC \cite{yao2015describing}    &  72.20\%     &  58.90\%    &  46.80\%    &  35.90\%    &  24.90\% \\ \hline
\hline
M$^3$-V    &  70.20\%     &  56.60\%    &  44.80\%    &  35.00\%   &  24.60\% \\
M$^3$-C    &  \bf{77.20\%}     &  \bf{61.30\%}    &  47.20\%    &  35.10\%   &  25.70\% \\
M$^3$-VC   &  73.60\%     &  59.30\%    &  \bf{48.26\%}    &  \bf{38.13\%}   &  \bf{26.58\%} \\

\hline
\hline
\end{tabular}
\vspace{3mm}
\caption{The performance comparison with SA \cite{yao2015describing} using different visual features on MSR-VTT. Here V, C, I and R denote VGG-19 \cite{simonyan2014very}, C3D \cite{tran2015learning}, Inception-V3 \cite{szegedy2015rethinking} and ResNet-50 \cite{he2015deep}, respectively.}
\label{msrvtt-all}
\end{table}
\subsubsection{Experimental Results on MSVD}
\begin{figure*}[t]
\begin{center}
\includegraphics[scale=0.5]{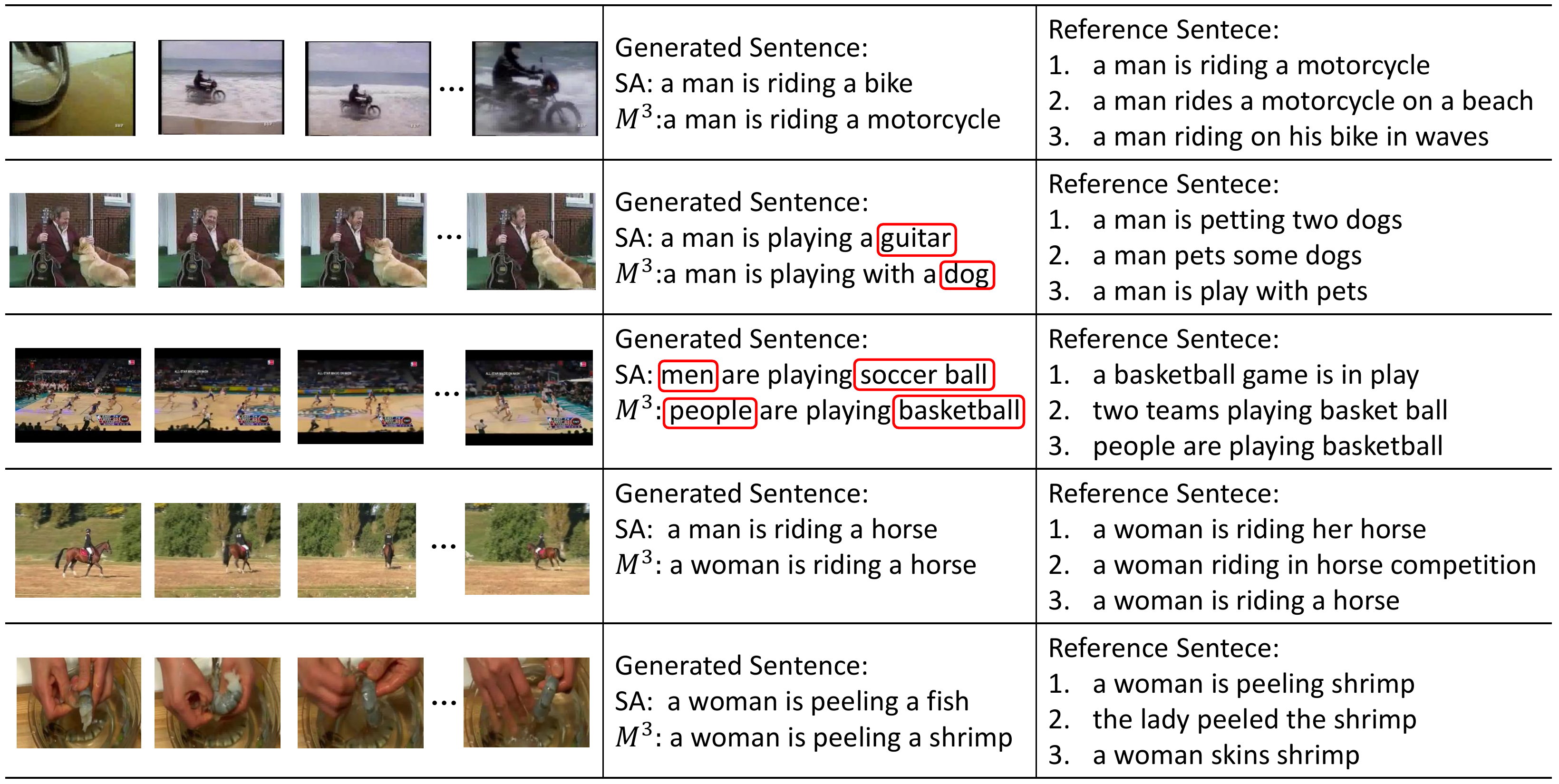}\\
\end{center}
\caption{Descriptions generated by SA-google, our M$^3$-google and human-annotated ground truth on the test set of MSVD. From these sentences in the second and third videos, M$^3$-google can generate more relevant object terms than SA-google (``basketball'' vs. ``soccer ball''), and M$^3$-google has a global visual attention on the targets (``dog''), which differs from SA-google's local visual attention on the non-targets (``guitar'').}
\label{figure:sample}
\end{figure*}
For comprehensive experiments, we evaluate and compare with the state-of-the-art methods using single visual feature and multiple visual feature fusion, respectively.

When using single visual feature, we evaluate and compare our model with the other eight state-of-the-art approaches(\cite{thomason2014integrating}, \cite{venugopalan2014translating}, \cite{yao2015describing}, \cite{venugopalan2015sequence}, \cite{pan2015jointly}, \cite{yu2015video}, \cite{pan2015hierarchical}, \cite{ballas2015delving}). The experimental results in terms of BLEU (n-gram) and METEOR are shown in Table \ref{msvd-rgb}. Here we give the best single feature results of the compared eight methods, and show the results of the proposed M$^3$ together with five single features, e.g., VGG-19 \cite{simonyan2014very}, C3D \cite{tran2015learning}, Inception-V3 \cite{szegedy2015rethinking}, ResNet-50 \cite{he2015deep} and GoogleNet \cite{szegedy2015going}. Among these compared methods, SA \cite{yao2015describing} is the most similar method to ours, which also has an attention-driven video encoder and LSTM-based text decoder but no external memory. When both models use the same GoogleNet feature, our M$^3$-google can make a great improvement over SA by $\frac{{51.17 - 40.3}}{{40.3}} = 26.9\% $ in the BLEU@4 score and by $\frac{{31.47 - 29.0}}{{29.0}} = 8.5\% $ in the METEOR score, respectively. It can be concluded that the better performance of our model benefits from multimodal memory modelling. In addition, our five M$^3$ models outperform all the other methods except HRNE \cite{pan2015hierarchical} in terms of METEOR. It is because HRNE \cite{pan2015hierarchical} specially focus on build a fine-grained video representation for captioning.
To further compare the results of the five M$^3$ models using different visual features, we can see that M$^3$-inv3 achieves the best performance, following by M$^3$-res, M$^3$-google and M$^3$-vgg19. The performance rank is very similar to that of these methods' image classification accuracy on ImageNet \cite{krizhevsky2012imagenet}, which proves that visual feature is very important for video captioning. Actually, the same conclusion has been drawn in image captioning where GoogleNet features obtain better results than VGG-19 features \cite{vinyals2016show}.

When using multiple visual feature fusion, we compare our model with the other four state-of-the-art approaches(\cite{yao2015describing}, \cite{venugopalan2015sequence}, \cite{pan2015jointly}, \cite{yu2015video}). The comparison results are shown in Table \ref{msvd-fuse}.
SA-G-3C \cite{yao2015describing} uses the combination of GoogleNet feature and 3D-CNN feature. S2VT-rgb-flow \cite{venugopalan2015sequence} uses the two-stream features consisting of RGB feature extracted from VGG-16 networks and optical flow feature extracted from AlexNet \cite{krizhevsky2012imagenet}. Both LSTM-E-VC \cite{pan2015jointly} and p-RNN-VC \cite{yu2015video} combine VGG-19 feature and C3D feature. We propose M$^3$-VC and M$^3$-IC for comparison. M$^3$-VC also uses VGG-19 feature and C3D feature while M$^3$-IC uses Inception-V3 feature and C3D feature. They all perform better than the other methods in terms of the two metrics, which proves the effectiveness of our model on long-term dependency modelling.

Fig. \ref{figure:sample} illustrates some descriptions generated by SA-google, our M$^3$-google and human-annotated ground truth on the test set of MSVD. From these sentences, we can see that both SA-google and M$^3$-google generate semantic relevant descriptions. However, it should be noted that our M$^3$-google can generate more relevant object terms than SA-google. For example, compared with the words ``men'' and ''soccer ball'' generated by SA-google in the third video, the words ``people'' and ``basketball'' generated by M$^3$-google are more precise to express the video content. Moreover, our M$^3$-google has a global visual attention on the targets, which differs from SA-google's local visual attention on the non-targets. For example, M$^3$-google predicts the target word ``dog'' in the second video while SA-google predicts the non-target word ``guitar''. All these results further demonstrate the effectiveness of our method.

\subsubsection{Experimental Results on MSR-VTT}

MSR-VTT is a recently released benchmark dataset \cite{Xu2016large} which has the largest number of video-sentence pairs. Considering that there are few methods tested on this dataset, we compare our model with SA \cite{yao2015describing} which is the most similar work to ours. Similarly, we perform experiments with these two methods using single visual feature and multiple visual feature fusion simultaneously. The comparison results are reported in Table \ref{msrvtt-all}. SA-V and SA-C use the VGG-19 feature and C3D feature, respectively. SA-VC fuses these two kinds of features. Our M$^3$-V, M$^3$-C and M$^3$-VC use the same features with the corresponding SA methods. It can be seen that our methods consistently outperform the corresponding SAs. The improved performance proves the importance of multimodal memory in our M$^3$ again.
In addition, from either M$^3$ or SA, we can see that the results from C3D feature are generally better than those using VGG-19 feature. It may be that the motion information is very critical for the video representation in this dataset, because C3D feature encodes both visual appearance and motion information in video.

\section{Conclusions and Future Work}

This paper proposes a Multimodal Memory Model (M$^3$) to describe videos, which builds a visual and textual shared memory to model the long-term visual-textual dependency and further guide global visual attention. The extensive experimental results on two publicly benchmark datasets demonstrate that our method outperforms the state-of-the-art methods in terms of BLEU and METEOR metrics.

As we can see from the experimental results, video representation is very important for the performance of  video captioning. In the future, we will consider to improve video representation learning algorithm, and integrate video feature extraction networks with multimodal memory networks to form an end-to-end deep learning system.

{\small
\bibliographystyle{ieee}
\bibliography{egbib}
}

\end{document}